\documentclass[conference]{IEEEtran}
\IEEEoverridecommandlockouts
\usepackage{cite}
\usepackage{amsmath,amssymb,amsfonts}
\usepackage{algorithmic}
\usepackage{graphicx}
\usepackage{textcomp}
\usepackage{booktabs}
\usepackage{xcolor}
\usepackage{multirow}
\usepackage{float}
\usepackage{hyperref}
\usepackage{tikz}
\usepackage{xcolor}

\usepackage[ruled,linesnumbered]{algorithm2e}

\DeclareMathOperator*{\argmax}{argmax} 

\begin{document}

\title{TriNE: Network Representation Learning for Tripartite Heterogeneous Networks\\
\thanks{This research is sponsored by U.S. National Science Foundation (NSF) through Grants IIS-1763452 \& CNS-1828181, and by Bidtellect Inc. through a sponsorship agreement.}
}

\author{\IEEEauthorblockN{Zhabiz Gharibshah and Xingquan Zhu}
\IEEEauthorblockA{\textit{Department of Computer \& Electrical Engineering and Computer Science, Florida Atlantic University}\\
Boca Raton, FL 33431, USA \\
\href{mailto:zgharibshah2017@fau.edu}{\{zgharibshah2017, xzhu3\}@fau.edu}
}}

\maketitle

\begin{abstract}
In this paper, we study network representation learning for tripartite heterogeneous networks which learns node representation features for  networks with three types of node entities. We argue that tripartite networks are common in real-world applications, and the essential challenge of the representation learning is the  heterogeneous relations between various node types and links in the network. 
To tackle the challenge, we develop a tripartite heterogeneous network embedding called TriNE.
The method considers unique user-item-tag tripartite relationships, 
to build an objective function to model explicit relationships between nodes (observed links), and also capture implicit relationships between tripartite nodes (unobserved links across tripartite node sets). The method organizes metapath-guided random walks to create heterogeneous neighborhood for all node types in the network. This information is then utilized to train a heterogeneous skip-gram model based on a joint optimization. Experiments on real-world tripartite networks validate the performance of TriNE for the online user response prediction using embedding node features. 
\end{abstract}


\section{Introduction}
Many real-world applications have tripartite relationships between different sets of entities. For example, e-commence systems usually involve user-product-vendor connections where users purchase products (user-product relations) and products are made by vectors (product-vendor relations). Similarly, health domains often have patient-disease-drug bonds, where patients are diagnosed with different diseases (patient-disease relations), and diseases can be treated by drugs (disease-drug relations). 

In a tripartite network, connections between entities are unique and have very interesting (and useful) meanings. For example, in a user-product-vendor tripartite network, the bipartite user-product and product-vendor relations are directly observable, whereas the user-vendor bipartite relationship is often unobservable which commonly refers to the loyalty of users to the brand~\cite{Zhang:2016:largescale}. On the other hand, in a patient-disease-drug tripartite network, the bipartite patient-disease and disease-drug relations are easy to collect, whereas the patient-drug bipartite relationship is often unknown and is commonly tied to adverse drug relations~\cite{Takarabe:2011:network}. Therefore, an accurate link prediction in a tripartite network setting is important to understand how users respond to the third party. 

Indeed, the unique three-party (\textit{i.e.}, tripartite) entity relationships provide useful information which is hard to be characterized by traditional heterogeneous information networks
. This also provides challenges for existing network embedding methods to be directly applied to the tripartite networks
\begin{itemize}
    \item \textbf{Imbalanced Nodes:} tripartite networks have multiple node types, and often have severely imbalanced nodes where one type of nodes are often much more than other types (\textit{e.g.} number of users are far more than the number of vendors). As a result, a traditional global random walk based network representation learning, such as node2vec~\cite{grover2016node2vec}, cannot learn effective features for tripartite network nodes.
    \item \textbf{Imbalanced network links:} In many tripartite networks, edges between two types of nodes, such as user-clicks on advertisements~\cite{Qu:2019:product} or user adverse drug reaction on particular drugs~\cite{Takarabe:2011:network} are rare events, which are making it difficult to learn effective features for the link prediction task.
\end{itemize}

While network embedding approaches, including bipartite network embeddings, have been commonly studied recently~\cite{network2020zhang}, no work currently exists to learn node representation for tripartite networks explicitly. Particularly in handling the above two challenges for tripartite network embedding learning.

Motivated by the above observations, in this paper, we propose TriNE, a tripartite network embedding learning algorithm to learn features to represent nodes for tripartite heterogeneous networks. 
TriNE considers coupled bipartite correlations to build an objective function to model explicit relationships between bipartite nodes (observed links between nodes), and also capture implicit relationships between tripartite nodes (unobserved links across tripartite node sets). The optimization allows the metapath-guided random walks carry out on the tripartite network to handle imbalanced node numbers for an effective link prediction. Experiments on real-world tripartite networks validate the performance of the proposed method.

\section{Related work}
Network representation learning has gained a lot of attentions recently~\cite{network2020zhang}. The classic dimension reduction methods like LLE\cite{Roweis2000}, ISOMAP\cite{tenenbaum_global_2000}, Laplacian eigenmaps\cite{Laplacian} and their extensions \cite{Manifold},\cite{GraphEmbedding} adopted various versions of linear and non-linear matrix factorization based techniques. Although the performance of these methods are shown successful for relatively small networks, but applying matrix decomposition is not scalable for large networks.

The network embedding has shown the decent performance in many data mining and recommendation systems. The initial studies followed the word2vec idea \cite{Mikolov:2013:Distributed} to extend the skip-gram models for homogeneous networks like LINE\cite{Tang2015LINELI}, node2vec\cite{grover2016node2vec}. These random walk based methods generate a corpus of random walk samples which are fed into skip-gram models to learn node embedding vectors.

Recently, several studies have been proposed for heterogeneous network embedding. Authors in \cite{Xu2017EmbeddingOE} proposed a method to encode inter- and intra-network edges in heterogeneous networks. In \cite{Dong2017metapath2vecSR}, authors extended DeepWalk\cite{perozzi2014deepwalk} method by introducing metapath-guided random walks and a heterogeneous skip-gram model. SHNE method which was proposed in \cite{SHNEZhang2019} investigated to integrate semantic information along with structural relations between nodes into the network embedding procedure.

There are many studies developed based on the architecture of graph neural networks. In \cite{MetapathFan2019}, authors proposed a metapath-based embedding method mainly designed using a graph neural network(GNN). The representation of the knowledge graph as the type of heterogeneous networks have received a lot of interest recently. Knowledge graphs (KG) are semantic heterogeneous networks including a collection of entities with attributes that are inter-connected together through edges. Here relations corresponding to edges may have different types and functionalities. They are usually described through a triplet like (Head, Relation, Tail) that a relation connects head and tail entities. This structure of data has been studied for different applications like the link prediction. 
Recently, a study \cite{Knowledge-AwareWang2019} elaborated a GNN-based method for node embedding in knowledge graphs in which weights of link between nodes like user and item in the network are not available. They suggested a supervised learning method to model a personalized scoring function to determine weights that followed a relational heterogeneity principle in the knowledge graph. In order to deal with data sparsity problem, they designed a leave-one-out loss function combined with a label smoothness regularization to predict the weight values of links. They then were used to calculate node embeddings through a local neighborhood aggregation.
In many knowledge graph based recommendation systems, the interactions  between two entities of user and item are usually modelled by applying aggregation mechanisms like average pooling and attention units over their user-defined neighborhood based on calculated node embedding vectors. Then, a sigmoid function with inner-product kernel is used to represent the probability of links\cite{Tang2015LINELI},\cite{DKNWang2018}. The authors in \cite{End-to-EndQu2019} proposed a neighborhood interaction model to integrate high order neighbor-neighbor interactions for training in a graph neural network.
They improved the performance of predicting user click-through rate values as the link between user and item entities using a bi-attention network in their design.

In \cite{RippleNet} authors suggested an end-to-end learning method called RippleNet to unify the user preference propagation within a knowledge graph embedding to calculate the distribution of user interests regarding an item.
In this paper, we intend to propose a network representation method to embed a tripartite heterogeneous network into a low-dimensional vector space.

\section{Preliminaries \& Problem Definition}

Let $\mathcal{G}=\{\mathcal{V},\mathcal{E}\}$ denotes a heterogeneous network (graph) which includes a vertex set ($\mathcal{V})$, an edge set $\mathcal{E}$ with different types. They are characterized with two type mapping functions as $\phi: \mathcal{V} \rightarrow O$ and $\psi: \mathcal{E} \rightarrow R$ where $O$ is the set of node types and $R$ is the set of edge types in the network. Each vertex(entity) $v \in \mathcal{V}$ in the graph is associated with vertex type using $\phi(v) \in O$. Similarly, each edge $e \in \mathcal{E}$ belongs to an edge type specified by $\psi(e) \in R$. A tripartite heterogeneous graph is the specific type of the heterogeneous graph where $|O|=3$ and $|R|=3$. The node set $\mathcal{V}$ consists of three node types $\mathcal{V}=\mathcal{V}_1 \cup \mathcal{V}_2 \cup \mathcal{V}_3$, which in turn encompasses three types of entities in the tripartite network, such as user nodes ($\mathcal{V}_1$), page categories ($\mathcal{V}_2$), and item category nodes ($\mathcal{V}_3$) in a tripartite user interaction network. Due to the tripartite network nature, we assume network edges always connect nodes between two different parties, \textit{i.e.} $\mathcal{E}={\mathcal{E}_1 \cup \mathcal{E}_2 \cup \mathcal{E}_3}$ where $E_1=(\mathcal{V}_1 \times \mathcal{V}_2)$, $\mathcal{E}_2=(\mathcal{V}_2 \times \mathcal{V}_3)$ and there is $\mathcal{E}_3=(\mathcal{V}_1 \times \mathcal{V}_3)$. In the case of recommendation systems and online advertising, the portion of links between $\mathcal{V}_1$ and $\mathcal{V}_3$ are missing. Therefore, the connectivity between users and items(Ads) is generally evaluated as the target \cite{DKNWang2018},\cite{DINZhou2018},\cite{DeepGharibshah2020}.

In the heterogeneous network, dealing with different types of nodes brings in a semantic relationship between nodes and edges. This complexity is generally described with a meta-data modeling through the network schema and metapaths. 

\begin{figure}[h]
  \centering
  \includegraphics[width=0.4\textwidth]{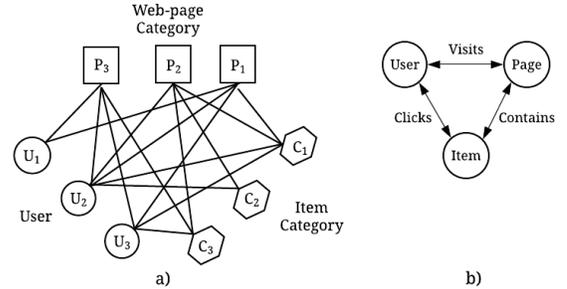}
    \caption{a) An example of heterogeneous network including three types of nodes and inter-connecting edges b) Network schema}
    \label{fig:HIN_example}
\end{figure}
Formally, metapath is defined as a semantic path in the heterogeneous graph. Following the network schema in this case, the path is shown as a sequence of different relations between node types as:
$A_0\xrightarrow{R_0}A_1\xrightarrow{R_1}A_2~...\xrightarrow{R_n}A_{n+1}$ where $R=R_0 R_1R_2...R_n$ describes a composite relation between node type $A_0$ and $A_{n+1}$.

Figure \ref{fig:HIN_example} shows an example of a tripartite network along with the corresponding network schema. Different metapaths represent different semantic information. For example, metapath "UPC" indicates a user visiting a page category which contains an item's category. Moreover, "UPCPU" represents two users who visit page categories including the same item category type. With regard to metapaths, metapath-guided random walks and metapath-guided neighbors concepts are also introduced~\cite{HeterogeneousShi2017},\cite{MetapathFan2019}. The metapath-guided random walk is considered as a random walk started from any node in the network which follows the specific metapath to randomly select the next node of the walk. The generation of the next node of the random walk is continuously repeated over the metapath scheme until reaching to the user-defined length. Metapath-guided neighbors are similarly defined as the  visited nodes when generating random walks followed by a given metapath. In this case, local neighborhood around nodes are defined according to the step travelled from a specific source node. So the neighbors of each node is considered as the combination of the visited nodes till given number of steps are taken. For example, in Figure \ref{fig:HIN_example}, if we denote i-step neighbors of the node $x$ guided by the metapath $\rho$ as $N^i_\rho(x)$, so   $\small{N^1_\rho(U_1)=\{P_3,P_1\}}$, $\small{N^2_\rho(U_1)=\{U_2,U_3,C_3,C_1\}}$. Then all metapath-guided neighbors of the node $U_1$ are $\small{N_\rho(U_1)=\{P_3,P_1,U_2,U_3,C_3,C_1\}}$. 
Several studies in literature showed that metapaths and extended versions of metapaths i.e meta-graphs can be seen as a tool to define the proximity and similarity measures\cite{MetaGraphZhao2017},\cite{PersonalizedYu2014},\cite{MetaGraph2VecZhang2018}. In this case, given a tripartite heterogeneous graph, the goal of problem in this paper is defined as the encoding of entities in the form of embedding vectors by using a proposed embedding approach. 

\begin{figure*}[h]
  \centering
  \includegraphics[width=0.8\textwidth]{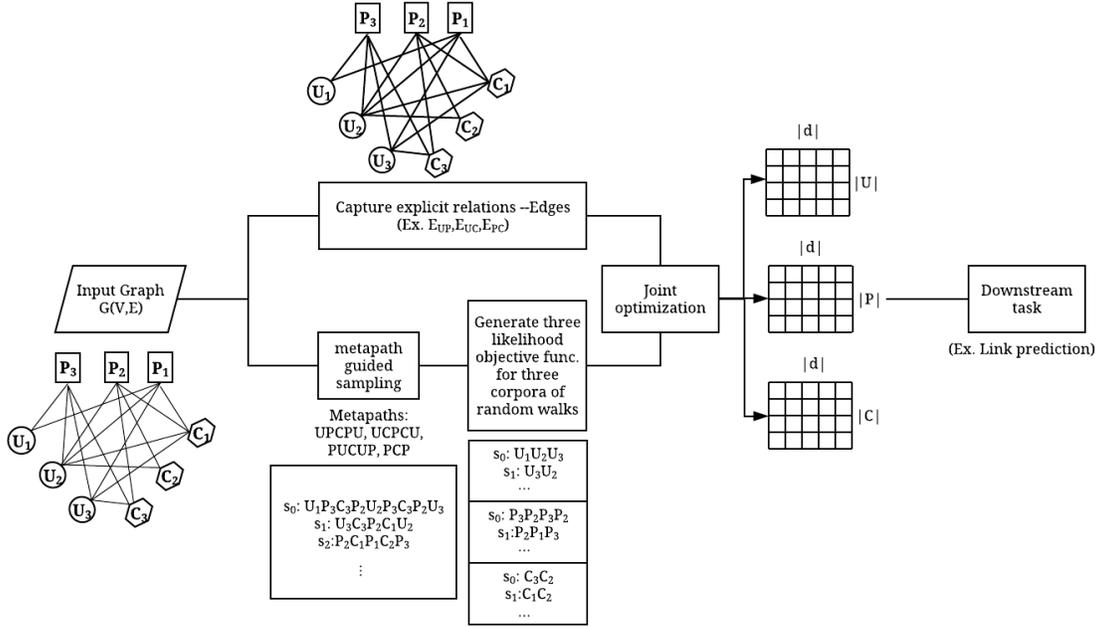}
    \caption{Tripartite network embedding structure. The algorithm includes two major components to capture explicit relations (edges) and implicit relations between nodes. The result is fused using a joint optimization to generate node embedding vectors.  
    }
    \label{fig:tripartite}
\end{figure*}



In the following section, we present a heterogeneous network embedding called TriNE for tripartite heterogeneous networks.

\subsection{Tripartite heterogeneous network embedding}

In order to learn the representation of nodes, algorithms like DeepWalk~\cite{perozzi2014deepwalk} and node2vec~\cite{grover2016node2vec} utilize the skip-gram model followed by random walks to map the concepts of words and context in word2vec~\cite{Mikolov:2013:Distributed} to networks. As those methods are designed to handle homogeneous networks, inspired by the recent progress on the network representation learning like BiNE~\cite{Gao:2018:BiNE} to deal with bipartite networks, we formulate the network representation learning method for tripartite heterogeneous networks  through the skip-gram model to account the implicit and explicit relations between nodes. 

Figure \ref{fig:tripartite} presents the overall workflow of our proposed method. We consider a tripartite heterogeneous graph as the input. The proposed method includes two major steps regarding capturing explicit and implicit information in the network. Explicit relations are observed edges between three types of node and implicit relations are transitive relations 
for different type nodes. They are characterized using metapath-guided neighbors gathered from metapath-guided random walks. In the embedding approach both explicit and implicit relations are addressed through a joint optimization. Given tripartite graph $\mathcal{G}=\{\mathcal{V},\mathcal{E}\}$, the goal is defined to learn a map function to project each node $v \in \mathcal{V}$ in tripartite network to a low-dimensional numerical vector $e_v \in \mathbb{R}^d$. 
It is expected that the learned embedding vectors can incorporate informative features to be used in the down-stream task like the predicting links in the network. 

\subsection{Implicit relation modeling}
Using the skip-gram architecture to model the network, the context of each node $v$ is defined as the set of surrounding nodes $N_c(v)$ connected to it through edges. For the embedding of each node in the homogeneous graph, the probability to see independently  each member of $N_c(v)$ in the embedded space can be defined as 
\begin{equation}
\small
\label{obj_function}
\begin{aligned}
P(N_c(v)\mid v) = \prod_{v_x\in N_c(v)} P(v_x\mid v;\theta)
\end{aligned}
\end{equation}

Therefore, the objective function is defined as a maximum likelihood problem to maximize the probability to observe the vector neighborhood as the context of each node in its feature representation. According to the three types of nodes in tripartite networks such as $\{\mathcal{V}_1,\mathcal{V}_2,\mathcal{V}_3\}$, three likelihood functions can be defined separately. For the all nodes within set $\mathcal{V}_1$, the first likelihood function is:
\begin{equation}
\small
\label{obj_function1}
\begin{aligned}
O_1 = \argmax_\theta \prod_{u_i\in \mathcal{V}_1}\prod_{u_x\in {{N_{c}}_{(\mathcal{V}_1)}}(u_i)} P(u_x\mid u_i;\theta) 
\end{aligned}
\end{equation}
where $\theta$ includes the embedding vectors of all nodes from the type of $\mathcal{V}_1$. ${N_{c}}_{(\mathcal{V}_1)}(u_i)$ also denotes all $\mathcal{V}_1$-type context nodes connected to the node $u_i$ via metapath-guided random neighbors in the network. In this case, $P(u_x\mid u_i;\theta)$ is usually approximated as a heterogeneous softmax function using the inner product kernel\cite{grover2016node2vec}, \textit{i.e.}:
\begin{equation}
\small
\label{softmax_function}
\begin{aligned}
P(u_x\mid u_i;\theta) = \dfrac{e^{\theta_{u_x}^T\theta_{u_i}}}{\sum_{u_k \in \mathcal{V}_1}{e^{\theta_{u_k}^T\theta_{u_i}}}}
\end{aligned}
\end{equation}


The maximization of the above optimization function is computationally expensive because all nodes are used for the calculation of the denominator in the softmax function. The general approach is to adopt a negative sampling which randomly samples relatively small set of nodes from some pre-defined distributions to approximate the denominator of softmax function. Here we employ a locality sensitive hashing method \cite{Gao:2018:BiNE} as follows: Given a node in the network, by defining the bucket of its neighbors based on the topological structure of tripartite network, negative samples are chosen from the remaining nodes different from the neighbors of the node. Therefore, instead of using softmax function to parameterize $P(u_x\mid u_i;\theta)$ in Eq. (2), the following equation is used:
\begin{equation}
\label{new_function}
\begin{aligned}
P(u_x\mid u_i;\theta) = \prod_{z\in {\{u_x\} \cup {N_{e}}_{(\mathcal{V}_1)}(u_i)}} P(z\mid u_i;\theta)
\end{aligned}
\end{equation}
where ${N_{e}}_{(\mathcal{V}_1)}(u_i)$ shows the negative samples of the node $u_i$ sampled among the $\mathcal{V}_1$ set and  $P(z\mid u_i;\theta)$ is also defined as: 

\begin{equation}
\label{new2_function}
\begin{aligned}
P(z\mid u_i;\theta)=
\begin{cases}
\sigma({\theta_{u_i}^T\theta_{z}}) & \text{if z is context of } u_i \\
1-\sigma({\theta_{u_i}^T\theta_{z}}) & \text{z} \in {N_{e}}_{(\mathcal{V}_1)}(u_i)
\end{cases}
\end{aligned}
\end{equation}
In the similar way, for each node $v\in\mathcal{V}_2$ and $w\in\mathcal{V}_3$, the optimization function can be shown as: 
\begin{equation}
\small
\label{obj_function2}
\begin{aligned}
O_2 = \argmax_\theta \prod_{v\in \mathcal{V}_2}\prod_{v_x\in {N_{c}}_{(\mathcal{V}_2)}(v)} P(v_x\mid v;\theta) 
\end{aligned}
\end{equation}
\begin{equation}
\small
\label{obj_function3}
\begin{aligned}
O_3 = \argmax_\theta \prod_{w\in \mathcal{V}_3}\prod_{w_x\in {{N_{c}}_{(\mathcal{V}_3)}}(w)} P(w_x\mid w;\theta) 
\end{aligned}
\end{equation}
where $P(v_x\mid v;\theta)$ and $P(w_x\mid w;\theta)$ are also parameterized by the function introduced in Eq. (\ref{new_function}) and (\ref{new2_function}):

\subsection{Tripartite Random Walk}
The procedure of optimization in our task includes the objective functions $O_1$, $O_2$ and $O_3$ each of which are based on node samples which can be visited in the network to build context nodes and negative samples. In order to transform the structure of network into the skip-gram model, we follow metapath-guided random walks to traverse node paths in the tripartite graph to incorporate meaningful semantic relations in sequence of nodes.

Here we perform a biased and flexible truncated random walk by using the  centrality measure calculated by HITS method \cite{Gao:2018:BiNE} to control the number of random walk per node for learning implicit relations between same type of nodes. At each step, the transition from each node to the next one is dependant to the type of node. We highlight the core procedure given a heterogeneous graph $\mathcal{G}=\{\mathcal{V},\mathcal{E}\}$ and the metapath scheme $\small{\rho=v_0\xrightarrow{R_0}v_1\xrightarrow{R_1}v_2~... \xrightarrow{R_t}v_t\xrightarrow{R_{t+1}}v_{t+1}~...\xrightarrow{R_n}v_{n+1}}$.
In each step, we select the next node using the following probability:

\begin{equation}
\footnotesize
  p(v_{i+1}|v_i,\rho) =
    \begin{cases}
       \dfrac{1}{|N^{(1)}_\rho(v_i)|} & \scriptsize{\text{if $(v_{i+1},v_i) \in N^{(1)}_\rho(v_i)$ and $\phi(v_{i+1})=t+1$  }}\\
      
      0 & \text{otherwise}
    \end{cases}       
    \label{metapath_randomwalk}
\end{equation}
where $N^{(1)}_\rho(v_i)$ is the first order neighbor of the node $v_i$ guided by the metapath of $\rho$, $v_i$ is t-type node and $v_{i+1}$ has the type of $t+1$ according to $\rho$. In tripartite network, there is no directed links within same type nodes, which means that the first order proximity between same type nodes is zero. Therefore, as it is shown in Figure \ref{fig:tripartite}, in the implicit relationship modeling, we generate a corpus of random walks guided by specified metapath schemes to incorporate the semantic information between nodes. Then, for each type of nodes, we check the random walks and filter out the nodes with different node types. In this case, we will have a sequence of nodes with same type to be used for skip-gram modeling. For each type nodes, we will optimize the loss functions $O_1$, $O_2$ and $O_3$.

\subsection{Explicit Relationship Modeling}
In previous sections, we discussed the approach to preserve the high order proximity within each set of node type through the metapath-guided random walk and the skip-gram model. However, the information conveyed by observed inter-connecting edges in the tripartite graph is also important to consider. The weight of links show the tie strength and the local first order proximity between two nodes in the network. Therefore, in order to model these observed relations, given $\mathcal{G}=\{\mathcal{V},\mathcal{E}\}$, the following joint probability between each pair of nodes ($u_i$,$v_j$), ($v_j$, $w_k$) and ($u_i$, $w_k$) can be defined where $\mathcal{V}=\{\mathcal{V}_1,\mathcal{V}_2,\mathcal{V}_3\}$, $u_i \in \mathcal{V}_1$, $v_j \in \mathcal{V}_2$, $w_k \in \mathcal{V}_3$ and $\mathcal{E}_1$ is the subset of links in the graph between the pair of $u_i$ and $v_j$, $\mathcal{E}_2$ denotes the edges starting from a node in $\mathcal{V}_2$ and ended to a node in $\mathcal{V}_3$ and $\mathcal{E}_3$ indicates the set of edges between pair of nodes from $\mathcal{V}_3$ and $\mathcal{V}_1$ node type sets respectively.

\begin{equation}
\small
\label{joint probability1}
\begin{aligned}
P_1(u_i,v_j) = \dfrac{w_{ij}}{\sum_{e_{ij} \in \mathcal{E}_1}w_{ij}}\\ P_2(v_j,w_k) = \dfrac{w_{jk}}{\sum_{e_{jk} \in \mathcal{E}_2}w_{jk}}\\ P_3(u_i,w_k) = \dfrac{w_{ik}}{\sum_{e_{ik} \in \mathcal{E}_3}w_{ik}}
\end{aligned}
\end{equation}
In this case, $w_{ij}$ and $w_{jk}$ are the weights of edge $e_{ij}$ and $e_{jk}$ between nodes in $\mathcal{V}_1$ and $\mathcal{V}_2$ or $\mathcal{V}_2$ and $\mathcal{V}_3$ parties respectively. To model these first order proximity information inspired by \cite{Gao:2018:BiNE}, we use the sigmoid function with the inner product kernel to estimate joint probability functions:
\begin{equation}
\small
\label{sig probability1}
\begin{aligned}
\hat{P_1}(u_i,v_j) = \dfrac{1}{1+e^{\mathbf{\theta}^\intercal_{u_i}\mathbf{\theta_{v_j}}}}\\
\hat{P_2}(v_j,w_k) = \dfrac{1}{1+e^{\mathbf{\theta}^\intercal_{v_j}\mathbf{\theta_{w_k}}}}\\
\hat{P_3}(u_i,w_k) = \dfrac{1}{1+e^{\mathbf{\theta}^\intercal_{u_i}\mathbf{\theta_{w_k}}}}
\end{aligned}
\end{equation}
where $\mathbf{\theta_{u_i}}$, $\mathbf{\theta_{v_j}}$ and $\mathbf{\theta_{w_k}}$ are the low-dimensional embedding vectors corresponding to nodes $u_i$, $v_j$ and $w_k$ in three types of nodes in the network. In order to preserve edge connection information, we define the objective function in form of KL-divergence \cite{Tang2015LINELI} to minimize the difference between the distribution of edges and the reconstructed distribution:
\begin{equation}
\small
\label{obj_func4}
\begin{aligned}
O_4 = KL(P_1||\hat{P_1})=\alpha_1- \sum\limits_{e_{jk} \in \mathcal{E}_1}w_{jk} \log{\hat{P_1}(u_i,v_j)}
\end{aligned}
\end{equation}
\begin{equation}
\small
\label{obj_func5}
\begin{aligned}
O_5 = KL(P_2||\hat{P_2})=\alpha_2- \sum\limits_{e_{jk} \in \mathcal{E}_2}w_{jk} \log{\hat{P_2}(v_j,w_k)}
\end{aligned}
\end{equation}
\begin{equation}
\small
\label{obj_func6}
\begin{aligned}
O_6 = KL(P_3||\hat{P_3})=\alpha_2- \sum\limits_{e_{ik} \in \mathcal{E}_3}w_{ik} \log{\hat{P_3}(u_i,w_k)}
\end{aligned}
\end{equation}
\subsection{Model Training}
To address implicit and explicit information in the tripartite network, we leverage the negative sampling strategy and metapath-guided random walks in addition to a reconstruction distribution function to define multiple objective functions. To preserve those information, the optimization step can be formed into a joint optimization framework through combining the objective functions as:

\begin{equation}
\small
\label{softmax_function3}
\begin{aligned}
\max \mathcal{L} = \alpha{_1}O_1+\alpha{_2}O_2+\alpha{_3}O_3+\beta{_1} O_4+\beta{_2}O_5+\beta{_3}O_6 
\end{aligned}
\end{equation}
where $\alpha_1$, $\alpha_2$, $\alpha_3$, $\beta_1$, $\beta_2$, and $\beta_3$ are hyper parameters. For optimization, similar to other embedding methods like node2vec, we first create three corpora of vertex sequences through metapath-guided random walks per node type. For each node, we set up context nodes and negative nodes with the same node type in the sampled node sequence. Then we utilize Stochastic Gradient Descent algorithm to optimize the joint model. To calculate the gradient step to update the node embedding and context vectors, we follow the formulations discussed in~\cite{Gao:2018:BiNE} to optimize the objective function. 
Therefore, the gradient of element $\beta{_1}O_4$ can be calculated as follows. It is applied also for $\beta{_1}O_5$ and $\beta{_1}O_6$ elements respectively:
\begin{equation}
\small
\label{grad_1}
\begin{aligned}
\vec{u_i}=\vec{u_i}-\eta(\gamma{w_{ij}}(1-\sigma(\vec{u}^T_i.\vec{v}_j)).\vec{v}_j)
\end{aligned}
\end{equation}
\begin{equation}
\small
\label{grad2}
\begin{aligned}
\vec{v}_i=\vec{v}_i-\eta(\gamma{w_{ij}}(1-\sigma(\vec{u}^T_i.\vec{v}_j)).\vec{u}_j)
\end{aligned}
\end{equation}
In the second step, we update node embedding and context vectors by calculating the gradient  $\alpha{_1}O_1$ element. We repeat this rule for two other objective function components; $\alpha{_2}O_2$ and $\alpha{_3}O_3$. The update rules are calculated as follows
\begin{equation}
\small
\label{grad3}
\begin{aligned}
\vec{u}_i=\vec{v}_i-\eta\{\sum_{\{v_c\}\cup  N^{ns}_S(v_j)}\alpha_1(I(z,u_i)-\sigma(\vec{u}^T_i.\vec{\theta}_z)).\vec{\theta}_z\}
\end{aligned}
\end{equation}
\begin{equation}
\small
\label{grad4}
\begin{aligned}
\vec{\theta}_z=\vec{\theta}_z-\eta\{\alpha_1(I(z,u_i)-\sigma(\vec{u}^T_i.\vec{\theta}_z)).\vec{u}_i\}
\end{aligned}
\end{equation}
Where $I(z,u_i)$ is defined as an indicator function to specify that if node $z$ is located in the context of node $u_i$ or not.
The pseudo code of training algorithm is listed in Algorithm \ref{alg_1}.

\subsection{TriNE Algorithm}
Algorithm~\ref{alg_1} lists the detailed procedure of the proposed tripartite network embedding method. It starts with initializing the node embedding vectors and context vectors. In the next step, the corpus of metapath-guided random walks are generated. They are split into three corpora with regard to node types to make homogeneous node sequences. Then, they are fed into the training procedure. The training iterations are repeated until the change between subsequent iterations become so small.   


\begin{algorithm}[!ht]
  \SetAlgoLined
  \KwData{$\mathcal{G}=\{\mathcal{V},\mathcal{E}\}$: a tripartite heterogeneous graph; metapath scheme set: $\rho= \{\rho_1,\rho_2,...\}$, embedding dimension size: d, number of negative samples: $ns$, minimal and maximal number of random walk per node: {\footnotesize{$minT, MaxT$}}, Neighborhood size: $k$, random walk length: $l$
  }
  \KwResult{The node embedding vectors: $\mathbf{U} \in \mathbb{R}^{|\mathcal{V}_1 \times d|}$, $\mathbf{P} \in \mathbb{R}^{|\mathcal{V}_2 \times d|}$, $\mathbf{C} \in \mathbb{R}^{|\mathcal{V}_3 \times d|}$}
Initialize the embedding vectors in $\mathbf{U}$, $\mathbf{P}$ and $\mathbf{C}$\;
Initialize the context vectors $\theta_{u_x}$, $\theta_{p_y}$,  $\theta_{c_z}$ in $\mathbf{U}$, $\mathbf{P}$ and $\mathbf{C}$;

\For{$v \in \mathcal{V}$}{
$w = max(H(v),minT,maxT); $ where $H(v)$ is the HITS centrality measure\;
  \For{$l$ = 1,$\cdots$,$w$}{
   Cur=MetapathRandomWalk($\mathcal{G},\rho,v,minT,maxT$)
   $S$+=Cur\;
  }
  }
  $S_U,S_P,S_C$=FilterNodeSequenceByType($S$)\;
\texttt{\\}
\For{$E = \mathcal{E}_1,\mathcal{E}_2,\mathcal{E}_3$}{
  \For{each edge $e \in E$}{
  $u,v \leftarrow e.u,e.v$\;
  update $\vec{u}$ and $\vec{v}$\ by Eq.\eqref{grad_1} and Eq.\eqref{grad2}\; 
  $S_1$,$S_2$= $S_U,S_P$~if~$e\in\mathcal{E}_1$~else~$S_P,S_C$~if~$e\in\mathcal{E}_2$~else~$S_U,S_C$\;
  \For{$v_i,v_c$ in the window of random walk $S_1$}{
  Do negative sampling\;
  Update $\vec{v_i}$ by Eq.\eqref{grad3}\;
  Update $\vec{\theta_z}$ by Eq.\eqref{grad4} where $z\in\{v_c\}\cup  N^{ns}_S(v_j)$\;
  }
  \For{$v_i,v_c$ in the window of random walk $S_2$}{
  Do negative sampling\;
  Update $\vec{v_i}$ by Eq.\eqref{grad3}\;
  Update $\vec{\theta_z}$ by Eq.\eqref{grad4} where $z\in\{v_c\}\cup  N^{ns}_S(v_j)$\;
  }  
  
  }
}
\texttt{\\}
\textbf{MetapathRandomWalk}($\mathcal{G}, \rho, v, minT, maxT$)\\
$s[0]\leftarrow v$\;
\For{$i= 1,\cdots,l-1$}{
  choose randomly $u$ using Eq.\eqref{metapath_randomwalk}\;
  s+=u\;
  }
return s\;
\caption{TriNE: Tripartite Network Embedding Learning}
\label{alg_1}
\end{algorithm}

\section{Experiments}
\subsection{Benchmark Datasets}
In our experiments, we use two tripartite heterogeneous networks prepared from two real-world datasets including VisualizeUs~\cite{nr} and MovieLens~\cite{nr}. The former contains records from the tagging behavior of users on photos in which the weighted edges denote the number of times to have tagging of user on a specific image. The Latter is one of widely movie recommendation dataset consisting of tagging records of various users with regard to the list of movie items. The summary of some basic statistics of prepared datasets is listed in Table \ref{Tb:datasets}.
\begin{table}[h]
	\centering
	\tabcolsep 1.75pt
	\caption{\small{Basic statistics of two datasets}}
	\renewcommand{\arraystretch}{1.2}
		\begin{tabular}{ccccc} 
 			\toprule[1pt]
			Dataset & \# of Users & \# of Tags & \# of Items & \# of Edges\\\hline
			VisualizeUs  & 3,911 & 21,076 & 5,013 & 46,546\\
			MovieLens & 58,834 & 8,704 & 2,462 & 660,800\\
 			\bottomrule[1pt]
	\end{tabular}
	\label{Tb:datasets}
\end{table}
\subsection{Experimental Settings}
As the common way to evaluate network embedding approaches, we choose the link prediction application for the evaluation task based on the intuition that the better representation of node in an embedding method can help us to predict links better in the network. 
Therefore, the link prediction can be cast as a binary classification problem to predict whether a user tags an image or a movie category. 
The link embedding vector samples are organized by calculating the average of embedding vectors of edge nodes. we randomly select one fifth of link samples as the test set while the remaining are the training set. We use the same embedding dimension as 128 to make a fair comparison between methods.
We compare our proposed method with two other methods, \textit{i.e.}, metapath2vec\cite{Dong2017metapath2vecSR} and BiNE\cite{Gao:2018:BiNE}. To evaluate the effectiveness of network embedding, 
we also consider the concatenation of 
embedding feature vectors of different methods for more comparisons. 
All experiments are evaluated based on the 
5-fold cross validation. After getting a representation of network embedding methods for user type nodes as the target in training and test set, we train a binary classifier using a MLP neural network(3 layers with 100 neurons) on the training set to predict the link on the test set. We compare the result with different classifiers, including Support vector Machines (SVM) with RBF kernel and logistic regression (LR).
We use the Area Under Receiver Operating Characteristics Curve (AUC-ROC) as the major evaluation metric because it shows the model accuracy of ranking positive cases versus negative ones. Besides, we also employ  the area under precision-recall curve (AUC-PR) along with the F1 score as the additional performance metrics.
\subsection{Baselines}
We compare the performance of the proposed method TriNE with following baseline methods.
\begin{itemize}
    \item\textbf{Metapath2vec:} The baseline representation learning method \cite{Dong2017metapath2vecSR} for heterogeneous networks which applies metapath-guided random walks to model a heterogeneous skip-gram model. 
    \item\textbf{BiNE:} The method \cite{Gao:2018:BiNE} develops embedding vectors for bipartite networks. In our experiment, as the original datasets are in the form of tripartite networks, we execute this method on the subset of bipartite network including desired node types for the link prediction task.
    \item\textbf{Metapath2vec+BiNE:} In this experiment, we combine the embedding vectors calculated by metapath2vec method \cite{Dong2017metapath2vecSR} with BiNE.
    \item\textbf{Metapath2vec+TriNE:} To evaluate the quality of features extracted from the proposed method, we concatenate the embedding vectors provided by the proposed method with those made by Metapath2vec method~\cite{Dong2017metapath2vecSR}.
\end{itemize}

\begin{table*}[!htb]
\centering
\caption{Average performance of the proposed method and the other baselines on two datasets}
\label{Res_bin}
\begin{tabular}{|l|l|l|l|l|l|l|l|} 
\hline
\multirow{2}{*}{Classifier} & \multirow{2}{*}{Algorithm} & \multicolumn{3}{c|}{\centering Visualize-US} & \multicolumn{3}{c|}{\centering MovieLens}  \\ 
\cline{3-8}
                            &                            & AUC-ROC & AUC-PR & F1                                     & AUC-ROC & AUC-PR & F1                                   \\ 
\hline
\multirow{5}{*}{LR}         & Metapath2vec               & 0.5807  & 0.1811 & 0.2165                                 & 0.6045  & 0.2683 & 0.2903                               \\ 
\cline{2-8}
                            & BiNE                       & 0.6053  & 0.2016 & 0.2718                                 & 0.6687  & 0.2833 & 0.3061                               \\ 
\cline{2-8}
                            & TriNE                      & 0.6311  & 0.2243 & 0.3009                                 & 0.6795  & 0.3795 & \textbf{0.4401}                               \\ 
\cline{2-8}
                            & Metapath2vec+BiNE          & \textbf{0.6675}  & 0.2569 & \textbf{0.3494}                                 & \textbf{0.7168}  & 0.3398 & 0.3799                               \\ 
\cline{2-8}
                            & Metapath2vec+TriNE         & 0.6185  & \textbf{0.2972} & 0.3220                                 & 0.6793  & \textbf{0.4208} & 0.4339                               \\ 
\hline
\multirow{5}{*}{MLP
  ~}    & Metapath2vec               & 0.7829  & 0.4398 & 0.4210                                 & 0.7366  & 0.3518 & 0.4052                               \\ 
\cline{2-8}
                            & BiNE                       & 0.6788  & 0.1942 & 0.2415                                 & 0.6994  & 0.3094 & 0.3939                               \\ 
\cline{2-8}
                            & TriNE                      & 0.7396  & 0.4314 & 0.4296                                 & 0.7146  & 0.3150 & 0.3475                               \\ 
\cline{2-8}
                            & Metapath2vec+BiNE          & \textbf{0.8424}  & 0.4353 & 0.3373                                 & 0.7417  & 0.3467 & 0.3299                               \\ 
\cline{2-8}
                            & Metapath2vec+TriNE         & 0.8289  & \textbf{0.6196} & \textbf{0.5254}                                 & \textbf{0.7452}  & \textbf{0.3920} & \textbf{0.4280}                               \\
\hline
\multirow{5}{*}{SVM (RBF)
  ~} & Metapath2vec               & 0.5631  & 0.3532 & 0.2153                                 & 0.5475  & 0.3776 & 0.1726                               \\ 
\cline{2-8}
                              & BiNE                       & 0.5113  & 0.1683 & 0.0442                                 & 0.5434  & 0.3170 & 0.1571                               \\ 
\cline{2-8}
                              & TriNE                      & \textbf{0.6585}  & \textbf{0.5319} & \textbf{0.4551}                                 & 0.5396  & 0.3909 & 0.1459                               \\ 
\cline{2-8}
                              & Metapath2vec+BiNE          & 0.5657  & 0.3542 & 0.2208                                 & \textbf{0.5632}  & 0.3951 & \textbf{0.2206}                               \\ 
\cline{2-8}
                              & Metapath2vec+TriNE         & 0.5575  & 0.3624 & 0.1957                                 & 0.5515  & \textbf{0.4215} & 0.1870                               \\
\hline
\end{tabular}

\end{table*}
\subsection{Performance Comparison}
Table \ref{Res_bin} reports the performance of different methods in two datasets.

{\bf TriNE outperforms the baselines}
In table \ref{Res_bin}, we show that by applying the proposed embedding methods we can 
perform better than other approaches. In the case of using the MLP classifier, The proposed embedding method TriNE combined metapath2vec shows the best performance in almost all cases.
It indicates the effectiveness of incorporating metapath semantic relations to learn node embedding vectors.

{\bf Hybrid embeddings is helpful.}
Comparing different methods, it can be seen that TriNE and also the hybrid method of metapath2vec and TriNE generally outperform the all baseline methods. We show that applying heterogeneous skip-gram models for all node types is more effective method 
for the link prediction. The variant of TriNE to combine embedded features with metapath2vec method outperform the remaining methods based on the majority of metrics. It presents the usefulness of metapath-guided random walks than typical random walks to capture the discriminative information for the link prediction.  
\def\BibTeX{{\rm B\kern-.05em{\sc i\kern-.025em b}\kern-.08emT\kern-.1667em\lower.7ex\hbox{E}\kern-.125emX}}
\section{Conclusion}
In this paper, we proposed a tripartite network embedding learning method to model explicit relationships between nodes (observed links between nodes), and also capture implicit relationships between tripartite nodes (unobserved links across tripartite node sets). We applied a joint optimization to train a heterogeneous skip-gram model to capture semantic and structural relations collected by utilizing metapath-guided random walks. We validated the performance of the proposed method on real-world data gathered for the link prediction task in the movie and image tagging. The results showed the effectiveness of the proposed method.

In our study, we applied a linear combination of multiple loss terms as the objective function to jointly model explicit and implicit relations in the tripartite heterogeneous graph for the link prediction task. In our future work, we are seeking to investigate a non-linear combination for the network representation learning in the other applicable tasks.

\bibliographystyle{IEEEtran}
\bibliography{ref}

\end{document}